\documentclass[letterpaper]{article} 
\usepackage[preprint]{aaai2027}  
\usepackage[hyphens]{url}  
\usepackage{graphicx} 
\usepackage{natbib}  
\usepackage{caption} 
\usepackage{algorithm}
\usepackage{algorithmic}

\usepackage{newfloat}
\usepackage{listings}
\DeclareCaptionStyle{ruled}{labelfont=normalfont,labelsep=colon,strut=off} 
\floatstyle{ruled}
\newfloat{listing}{tb}{lst}{}
\floatname{listing}{Listing}

\usepackage{booktabs}

\usepackage{amsmath}
\usepackage{subcaption}
\usepackage{tikz}

\title{Beyond Answers, Traces, and Efficiency: Evidence for Adaptive Reasoning}
\author {
    Sejin Kim\textsuperscript{\rm 1},
    Sundong Kim\textsuperscript{\rm 2},
}
\affiliations {
    \textsuperscript{\rm 1}Chungnam National University\\
    \textsuperscript{\rm 2}Gwangju Institute of Science and Technology\\
    sejin.kim@cnu.ac.kr, sundong@gist.ac.kr
}

\begin{document}

\maketitle

\begin{abstract}
Current AI systems are evaluated as reasoners through final-answer accuracy, reasoning traces, process supervision, and computational or human-relative efficiency.
Yet these measures do not establish whether reasoning adapts when tasks or evidence change.
We distinguish \emph{across-task adaptivity}, where acquired abstractions transfer or recombine across novel problems, from \emph{within-task adaptivity}, where hypotheses and actions are revised in response to new observations, feedback, or counterexamples.
Using ARC-AGI as a diagnostic case study rather than an alignment benchmark, we examine what existing measures leave unresolved when test-time search, sampling, or adaptation affects performance.
We propose intervention-oriented evaluation that tests abstraction transfer and hypothesis revision under controlled changes to tasks and evidence, with explicit reporting of test-time procedures and budgets.
This perspective supports evaluating systems for robustness, responsiveness to corrective feedback, and auditability under changing conditions.
\end{abstract}
\section{Introduction}

Reasoning evaluation has traditionally focused on outcome-based measures such as final-answer accuracy and pass rates~\cite{uesato2022outcome,lee2025reasoning}.
As reasoning systems have become more complex, evaluation has expanded to intermediate reasoning traces, process-level supervision, inference-time computation, and human-relative efficiency.
Chain-of-thought (CoT), process reward models (PRMs), and metrics such as relative human action efficiency (RHAE) exemplify this shift beyond answer correctness~\cite{wei2022cot,lightman2024supervision,chollet2026arc3}.
Together, these developments reflect a growing recognition that final outcomes provide limited evidence about how a solution was obtained.

Yet an important question remains: \emph{does the system change its reasoning appropriately when the task or available evidence changes?}
A correct answer does not establish that an acquired abstraction will transfer to a novel problem.
A plausible reasoning trace does not establish that the reasoning will change when new evidence contradicts an earlier hypothesis.
Likewise, efficient behavior does not reveal what the system learned or whether it can reuse that knowledge.
Outcome quality, trace quality, and efficiency therefore provide useful but incomplete evidence of adaptive reasoning.

This gap becomes more consequential when systems rely on test-time search, sampling, refinement, or interaction~\cite{wang2023selfconsistency,snell2025ttc,jeon2026neuralmcts}.
Such procedures can be legitimate components of reasoning, but higher performance alone does not reveal whether improvement comes from abstraction transfer, hypothesis revision, or simply more opportunities for selection.
This makes explicit test-time reporting and stronger behavioral evidence important for interpreting reasoning claims.

Recent work has framed LLM reasoning in terms of System-2 thinking~\cite{zhang2026system,kim2024sys2reason}.
We argue that adaptive reasoning requires evidence at two complementary scales.
\emph{Across-task adaptivity} concerns whether acquired abstractions can be transferred, reused, or recombined across novel problems.
\emph{Within-task adaptivity} concerns whether hypotheses and decisions are revised as observations, feedback, or counterexamples become available.
This distinction shifts evaluation from observing reasoning to testing how behavior responds to controlled changes in tasks and evidence.
Stronger evidence for adaptive reasoning therefore comes from behavior under intervention, rather than from a successful trajectory alone.

Adaptive reasoning is not equivalent to alignment, but both adaptive capabilities matter for systems operating under novelty, ambiguity, and human feedback.
Such systems must often reuse relevant knowledge while revising assumptions that no longer fit the available evidence.
Evidence of transfer and revision can therefore inform alignment-relevant properties such as robustness, responsiveness to corrective feedback, and auditability.

This paper makes three contributions.
First, we organize existing reasoning evaluations by the evidence they provide about outcomes, reasoning traces and processes, resource use, and human-relative efficiency.
Second, we characterize adaptive reasoning through across-task transfer and within-task revision, and argue for evaluating both through controlled changes to tasks and evidence.
Third, using ARC-AGI as a diagnostic case study, we outline an intervention-oriented evaluation perspective that explicitly reports test-time procedures and computational budgets.

\newpage
\section{What Existing Reasoning Evaluations Tell Us}

Reasoning evaluations provide different kinds of evidence about model behavior.
Our aim is not to rank or replace existing metrics, but to clarify what each establishes and what it leaves unresolved.
These forms of evidence address related but different questions: whether a task was solved, what reasoning was expressed, whether intermediate steps were valid, what resources were required, and how efficiently behavior compared with humans.
We view them as complementary forms, each supporting a different class of reasoning claims rather than progressively sufficient definitions of reasoning.

\textbf{Outcome-based evidence.}
Accuracy, Pass@$k$, and success rate directly measure whether a system solves a task and remain fundamental indicators of task competence~\cite{uesato2022outcome,lee2025reasoning}.
Their limitation is not that they are uninformative, but that the same successful outcome can arise from substantially different procedures~\cite{hwang2025solution}.
A correct solution may reflect a transferable abstraction, a memorized pattern, a brittle heuristic, extensive search, or large-scale candidate sampling.
This outcome equivalence makes final performance insufficient for distinguishing how a solution was obtained or what knowledge would persist beyond the evaluated instance.
Outcome metrics therefore establish task success, but do not by themselves establish the reasoning process responsible for it.

\textbf{Trace-based evidence.}
Reasoning traces such as chain-of-thought, natural-language rationales, scratchpads, or intermediate programs expose information that final answers omit~\cite{wei2022cot,lee2025reasoning}.
They can reveal candidate hypotheses, decompositions, and intermediate representations, making a solution more inspectable.
However, an observed trace is an artifact of reasoning, not by itself evidence that the underlying behavior is adaptive.
A rationale may be weakly coupled to the decision that produced the answer~\cite{turpin2023faithfulness}, and observing a successful trace does not establish how the system would respond if relevant evidence changed.
Moreover, even a faithful trace answers only part of the question.
Faithfulness concerns whether the trace reflects the process associated with a decision, whereas adaptivity concerns whether that process changes appropriately under different evidence.

\textbf{Process-based evidence.}
Process-level evaluation goes further by assessing intermediate reasoning steps rather than merely exposing them.
Process reward models (PRMs), for example, can evaluate whether individual steps are valid or useful and provide denser supervision than outcome-only rewards~\cite{uesato2022outcome,lightman2024supervision}.
This is an important advance, but step quality and evidence responsiveness are distinct evaluation axes.
A trajectory may consist of locally valid steps without establishing whether the system would revise an earlier hypothesis when subsequent observations contradict it.
A high process score can therefore support a claim about the quality of an observed trajectory without establishing its behavior under changed evidence.
Process evaluation asks whether intermediate reasoning is valid; adaptive evaluation additionally asks whether that reasoning updates appropriately when its evidential basis changes.

\newpage
\textbf{Resource-based evidence.}
The growing use of test-time search, sampling, refinement, and additional inference compute introduces another dimension to reasoning evaluation~\cite{wang2023selfconsistency,brown2024monkeys,snell2025ttc,jeon2026neuralmcts}.
Systems achieving similar accuracy may require substantially different numbers of samples, tokens, search steps, refinement rounds, or interactions.
Reporting these quantities is important because it separates achieved performance from the resources used to obtain it.
Additional computation is not inherently undesirable: search, iteration, and refinement can themselves be components of reasoning~\cite{pourcel2025self}.
The unresolved issue is interpretive.
An accuracy gain may result from a better abstraction or more effective hypothesis revision, but it may also result from a larger opportunity to generate, search, and select candidates.
Resource reporting makes this distinction more visible without resolving it on its own.

\textbf{Human-relative evidence.}
Human-centric metrics add a behavioral reference point by comparing how efficiently humans and AI systems solve the same task.
Relative human action efficiency (RHAE), used in ARC-AGI-3, captures differences that final accuracy cannot: a human and a model may both succeed while requiring substantially different numbers of actions~\cite{chollet2026arc3}.
Such comparisons provide useful evidence about behavioral efficiency and can expose gaps hidden by binary success measures.
Human trajectories can provide still richer references, including common action sequences, strategy distributions, and points at which behavior changes~\cite{kim2026arctraj,legris2025harc}.
However, human-like efficiency should not be treated as a normative definition of good reasoning, nor does trajectory similarity establish that humans and models acquired the same abstractions or revised the same hypotheses~\cite{kim2025intention}.
Human-relative evidence therefore characterizes behavioral efficiency and trajectory patterns, while leaving the underlying adaptive capability unresolved.

Tab.~\ref{tab:evidence} summarizes these complementary forms.
Each expands the evidence available for reasoning claims, but none directly tests whether reasoning itself changes appropriately when relevant task properties or available evidence change.
This remaining question motivates our account of adaptive reasoning in the next section.

\begin{table}[t]
\centering
\small
\begin{tabular}{p{0.25\columnwidth} p{0.65\columnwidth}}
\toprule
\textbf{Evidence} & \textbf{What it establishes / leaves unresolved} \\
\midrule

Outcome &
Measures task success; does not identify how the solution was obtained. \\

Trace &
Exposes expressed reasoning; does not establish behavioral grounding or adaptation. \\

Process &
Evaluates intermediate-step quality; does not test responsiveness to changed evidence. \\

Resource &
Reports compute and search budgets; does not separate stronger reasoning from larger search. \\

Human-relative &
Measures efficiency against humans; does not identify what was acquired or revised. \\

\bottomrule
\end{tabular}
\caption{Complementary evidence provided by existing reasoning evaluations and the questions each leaves unresolved.}
\label{tab:evidence}
\end{table}
\begin{figure}[t]
\centering

\begingroup

\begin{minipage}[t]{0.44\linewidth}
\centering

\parbox[c][1.80cm][c]{\linewidth}{%
\centering
\small
Final Answer\\
Reasoning Trace\\
Step Quality\\
Resource Use / Efficiency\\
Human-Relative Evidence
}

\vspace{-1pt}

$\underbrace{\hspace{0.94\linewidth}}_
{\mbox{\scriptsize What was observed?}}$

\end{minipage}
\hfill
%
\begin{minipage}[t]{0.07\linewidth}
\centering

\parbox[c][1.80cm][c]{\linewidth}{%
\centering
$\longrightarrow$
}

\end{minipage}
\hfill
%
\begin{minipage}[t]{0.43\linewidth}
\centering

\parbox[c][1.80cm][c]{\linewidth}{%
\centering
\small
Behavior Under Intervention
}

\vspace{-1pt}

$\underbrace{\hspace{0.94\linewidth}}_
{\mbox{\scriptsize How does it change?}}$

\end{minipage}

\endgroup

\caption{Existing evidence characterizes what was observed; intervention-oriented evaluation tests how behavior changes under controlled changes to tasks or evidence.}
\label{fig:evidence-layers}
\end{figure}

\section{What Is Missing? Evidence for Adaptivity}

The evidence forms above characterize behavior on a particular task or trajectory, but claims of adaptive reasoning require evidence that controlled changes in relevant task conditions or available evidence induce systematic, corresponding behavioral changes.
We treat \emph{adaptivity} as a behavioral property rather than a claim about a particular internal mechanism.
We distinguish two complementary scales: \emph{across-task adaptivity} concerns whether acquired abstractions transfer or recombine across changing tasks, whereas \emph{within-task adaptivity} concerns whether hypotheses and subsequent decisions are revised as task-specific evidence changes.

\textbf{Behavioral notation.}
Let $\tau$ denote a task, $E_t$ the evidence available by step $t$, with $E_0$ denoting the initial task evidence, and $q$ a fixed test-time budget.
We write $B_\theta(\tau,E_t;q)$ for the observable behavior of a system $\theta$ under these conditions.
Let $C(\tau)$ denote the evaluator-specified task structure relevant to the reasoning claim.
It specifies what a controlled task variation is intended to preserve or change; it need not correspond to an accessible internal representation of the model.

\textbf{Across-task adaptivity: transferring and recombining abstractions.}
For across-task adaptivity, consider a controlled intervention $I_T$ producing $\tau' = I_T(\tau)$.
We define
\[
\Delta C = C(\tau') \ominus C(\tau),
\;\;
\Delta B =
B_\theta(\tau',E_0;q)
\ominus
B_\theta(\tau,E_0;q),
\]
where $\ominus$ denotes an evaluator-specified, task-appropriate contrast between structures or behaviors.
The central evidential relation is
\[
\fbox{$\Delta C \sim \Delta B$},
\]
where $\sim$ denotes an evaluator-defined consistency relation between the task change and the behavioral response.
It is not a universal similarity metric, but specifies what behavior should remain stable and what should change under the intervention.

This formulation makes across-task adaptivity more specific than generic generalization.
Standard generalization asks whether a model performs well on unseen examples.
A model can satisfy this criterion while relying on correlations unrelated to the abstraction implied by a reasoning claim.
Across-task adaptivity instead asks whether behavior tracks the structure deliberately preserved, recombined, or altered across related tasks.

A useful distinction is between \emph{concept-preserving} and \emph{concept-changing} interventions.
In a concept-preserving intervention, surface properties such as object identities, layouts, ordering, or irrelevant context may change while the relation required for the solution remains fixed.
Here, $C(\tau') = C(\tau)$ with respect to the target abstraction, so the corresponding behavior should remain stable or transform predictably.
In a concept-changing intervention, the relevant structure itself changes, and adaptive behavior should change consistently with that modification.
Evidence for transfer therefore combines stability to changes that should not matter with sensitivity to changes that should.

This criterion does not require identical reasoning traces or action sequences across tasks.
The same abstraction may support different strategies in different contexts.
Nor does it require accuracy to remain constant under every variation.
A harder recombination may reduce performance while still revealing reuse of the relevant structure.
The evidential target is the dependency between controlled structural variation and corresponding behavior, rather than performance invariance alone.

\textbf{Within-task adaptivity: revising hypotheses as evidence changes.}
Early observations may support several interpretations, while later counterexamples, feedback, or interaction outcomes may invalidate an intermediate hypothesis.
Prior work has explored iterative refinement and hypothesis revision using such signals~\cite{madaan2023selfrefine,shinn2023reflexion,wang2024hypothesissearch}.
Within-task adaptivity asks whether subsequent reasoning changes specifically in response to the evidence rather than merely persisting with an initial interpretation.

Let $I_E$ denote an evidence intervention at step $t$, producing
\[
E_t' = I_E(E_t),
\qquad
\Delta E = E_t' \ominus E_t.
\]
Because an evaluator may not observe an explicit internal hypothesis, let $Z_\theta(\tau,E_t;q)$ denote an observable intermediate decision, such as a hypothesis, proposal, program, action plan, or tool call.
The relevant dependency is
\[
\fbox{$
\Delta E
\;\longrightarrow\;
\Delta Z_t
\;\longrightarrow\;
\Delta B_{>t}
$},
\]
where $\Delta Z_t$ denotes the corresponding change in the observable intermediate decision and $\Delta B_{>t}$ the subsequent behavioral change.
The revision should be systematically consistent with what the new evidence supports or rules out.

Behavioral change alone is insufficient.
Different outputs can result from stochastic sampling, repeated retries, or unrelated variation.
If new evidence contradicts one component of a hypothesis, stronger evidence of revision is obtained when the corresponding component of $Z_t$ changes rather than when the system changes arbitrarily.
Conversely, evidence consistent with the current hypothesis should not induce unrelated revision.
The evaluation need only establish a meaningful relationship between changes in evidence and changes in reasoning, without prescribing a unique update.

This distinction is particularly important for separating revision from repeated search.
Two systems may initially fail and later produce the same correct answer.
One may incorporate information about why an earlier proposal failed and modify the relevant part of its next proposal.
Another may generate additional independent candidates until one succeeds.
Both are legitimate inference strategies, but they support different claims.
The former provides direct behavioral evidence of evidence-sensitive revision, whereas the latter primarily demonstrates the effectiveness of additional sampling or selection.
Holding the test-time budget $q$ explicit helps prevent these differences from being obscured by unequal search opportunities.

\textbf{Why are both scales needed?}
Across-task and within-task adaptivity capture related but non-interchangeable properties.
A system may transfer a useful abstraction across problems yet fail to revise it when contradictory evidence appears within a task.
Conversely, a system may respond effectively to local feedback while relying on representations that do not transfer beyond the current instance.
Together, the two scales examine whether reasoning can both \emph{reuse} knowledge across changing contexts and \emph{revise} it when the available evidence no longer supports it.

\textbf{What counts as adaptive evidence?}
Across both scales, three properties strengthen an adaptivity claim.
First, \emph{relevance}: behavior should respond to changes that matter to the current abstraction or hypothesis while remaining relatively stable to irrelevant variation.
Second, \emph{specificity}: the component of reasoning that changes should correspond to the information altered by the intervention rather than reflecting arbitrary global change.
Third, \emph{consistency}: comparable interventions should produce comparable response patterns across instances rather than isolated stochastic changes.
These criteria are not intended as a universal scalar metric; they specify the behavioral relationship that an evaluation should expose.

\textbf{Observation is not intervention.}
Outcome, trace, process, resource, and human-relative evaluations primarily characterize behavior that has already occurred.
Controlled variants have also been used to probe model behavior under targeted changes in prior evaluation work~\cite{gardner2020contrastsets,ribeiro2020checklists}.
Here, intervention-oriented evidence changes a variable that the claimed reasoning should depend on and measures the corresponding response.
Across tasks, this can mean preserving or modifying a relevant structure; within a task, it can mean adding an observation, counterexample, constraint, or feedback signal that changes support for a current hypothesis.
The result is a paired comparison between a change in relevant information and a subsequent behavioral response.

Instead of asking only \emph{what reasoning was produced}, adaptive evaluation asks \emph{how reasoning changes when relevant evidence changes}.
Fig.~\ref{fig:evidence-layers} summarizes how this perspective extends existing forms of reasoning evidence.

These forms of evidence remain complementary, while controlled intervention adds distinct evidence that directly probes the transfer and revision dependencies defined above.

Behavioral interventions do not identify a model's complete internal causal mechanism.
Different mechanisms may produce similar responses, and determining whether an intervention cleanly preserves or alters only the intended structure can itself be difficult.
Our claim is narrower.
Controlled behavioral variation provides stronger evidence for abstraction transfer and hypothesis revision than observing a successful trajectory in isolation.
We treat across-task and within-task adaptivity as a minimal diagnostic framework for adaptive reasoning claims, not as a complete definition of reasoning.
\section{ARC-AGI as a Diagnostic Case Study}

\begin{figure*}[t]
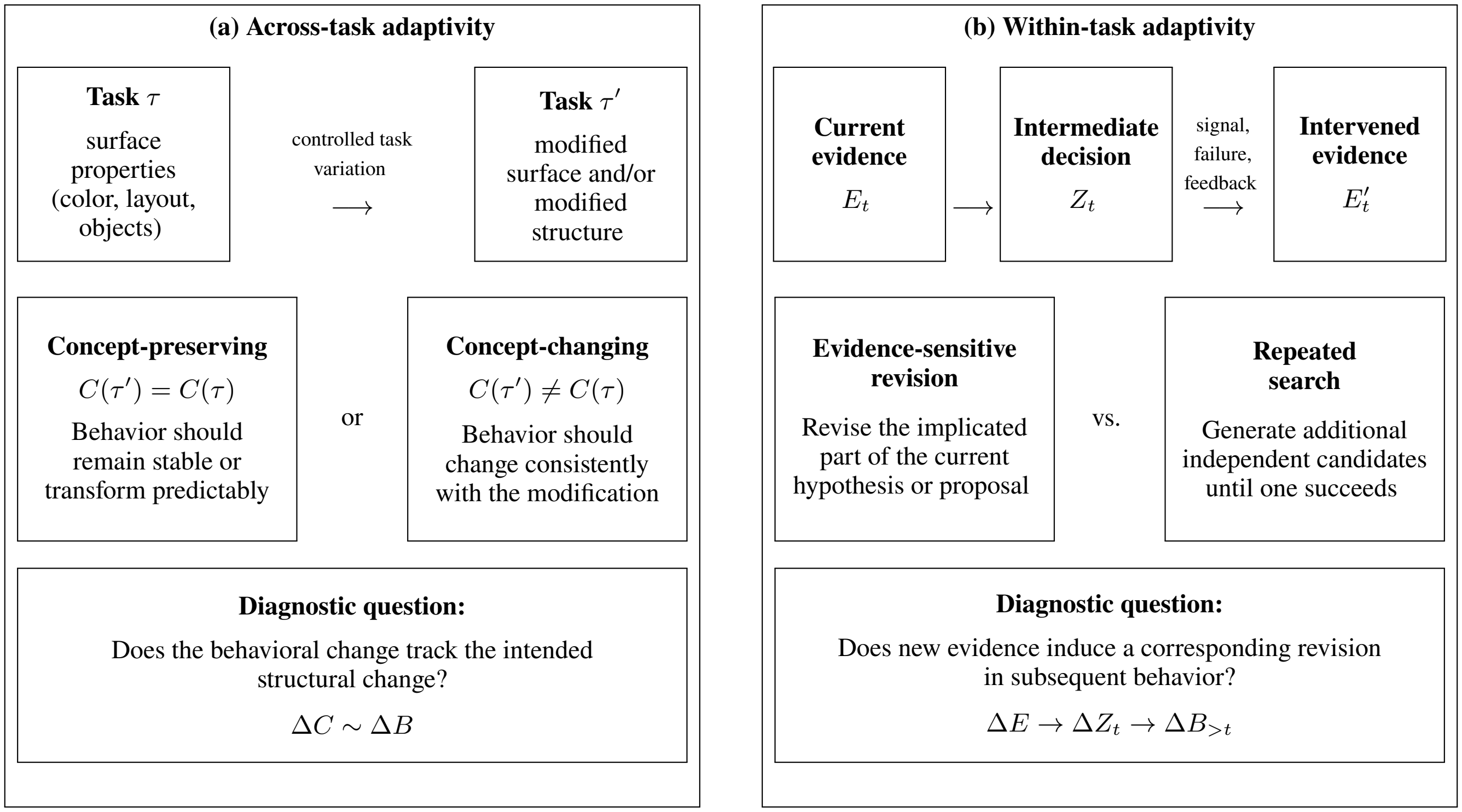

\centering

\begingroup
\setlength{\fboxsep}{4pt}
\setlength{\fboxrule}{0.4pt}


\newcommand{\diagbox}[3]{%
\fbox{%
\parbox[c][#2][c]{#1}{%
\centering
\footnotesize
#3
}}}

\newcommand{\arrowblock}[2]{%
\parbox[c][2.05cm][c]{#1}{%
\centering
\parbox[c][0.95cm][c]{\linewidth}{%
\centering
\scriptsize
#2
}%
\par\vspace{3pt}
$\longrightarrow$
}}

\newcommand{\acasebox}[4]{%
\diagbox{#1}{2.65cm}{%
\parbox[c][0.40cm][c]{0.94\linewidth}{%
\centering
\textbf{#2}
}%
\par\vspace{3pt}
\parbox[c][0.35cm][c]{0.94\linewidth}{%
\centering
#3
}%
\par\vspace{4pt}
\parbox[c][1.05cm][c]{0.94\linewidth}{%
\centering
#4
}%
}}

\newcommand{\bcasebox}[3]{%
\diagbox{#1}{2.65cm}{%
\parbox[c][0.62cm][c]{0.94\linewidth}{%
\centering
\textbf{#2}
}%
\par\vspace{5pt}
\parbox[c][1.45cm][c]{0.94\linewidth}{%
\centering
#3
}%
}}

\begin{minipage}[t]{0.485\textwidth}
\centering

\fbox{%
\parbox[t][9.4cm][t]{0.94\linewidth}{%
\centering

\small\textbf{(a) Across-task adaptivity}

\vspace{8pt}

\makebox[\linewidth][c]{%
\diagbox{0.28\linewidth}{2.05cm}{%
\textbf{Task $\tau$}\\[5pt]
surface\\
properties\\
(color, layout,\\
objects)
}%
\hfill
\arrowblock{0.24\linewidth}{%
controlled task\\
variation
}%
\hfill
\diagbox{0.28\linewidth}{2.05cm}{%
\textbf{Task $\tau'$}\\[5pt]
modified\\
surface and/or\\
modified\\
structure
}%
}

\vspace{11pt}

\makebox[\linewidth][c]{%
\acasebox{0.38\linewidth}%
{Concept-preserving}%
{$C(\tau') = C(\tau)$}%
{Behavior should\\
remain stable or\\
transform predictably}%
\hfill
\parbox[c][2.65cm][c]{0.06\linewidth}{%
\centering
or
}%
\hfill
\acasebox{0.38\linewidth}%
{Concept-changing}%
{$C(\tau') \neq C(\tau)$}%
{Behavior should\\
change consistently\\
with the modification}%
}

\vspace{8pt}

\makebox[\linewidth][c]{%
\fbox{%
\parbox[c][2.05cm][c]{%
\dimexpr\linewidth-2\fboxsep-2\fboxrule\relax}{%
\centering
\footnotesize
\textbf{Diagnostic question:}\\[5pt]
Does the behavioral change track the intended\\
structural change?\\[6pt]
$\Delta C \sim \Delta B$
}}}

}} 

\end{minipage}%
\hfill%
\begin{minipage}[t]{0.485\textwidth}
\centering

\fbox{%
\parbox[t][9.4cm][t]{0.94\linewidth}{%
\centering

\small\textbf{(b) Within-task adaptivity}

\vspace{8pt}

\makebox[\linewidth][c]{%
\diagbox{0.22\linewidth}{2.05cm}{%
\textbf{Current}\\
\textbf{evidence}\\[5pt]
$E_t$
}%
\hfill
\arrowblock{0.08\linewidth}{}%
\hfill
\diagbox{0.22\linewidth}{2.05cm}{%
\textbf{Intermediate}\\
\textbf{decision}\\[5pt]
$Z_t$
}%
\hfill
\arrowblock{0.15\linewidth}{%
signal,\\
failure,\\
feedback
}%
\hfill
\diagbox{0.22\linewidth}{2.05cm}{%
\textbf{Intervened}\\
\textbf{evidence}\\[5pt]
$E_t'$
}%
}

\vspace{11pt}

\makebox[\linewidth][c]{%
\diagbox{0.38\linewidth}{2.65cm}{%
\textbf{Evidence-sensitive}\\
\textbf{revision}\\[7pt]
Revise the implicated\\
part of the current\\
hypothesis or proposal
}%
\hfill
vs.
\hfill
\diagbox{0.38\linewidth}{2.65cm}{%
\textbf{Repeated}\\
\textbf{search}\\[7pt]
Generate additional\\
independent candidates\\
until one succeeds
}%
}

\vspace{8pt}

\makebox[\linewidth][c]{%
\fbox{%
\parbox[c][2.05cm][c]{%
\dimexpr\linewidth-2\fboxsep-2\fboxrule\relax}{%
\centering
\footnotesize
\textbf{Diagnostic question:}\\[5pt]
Does new evidence induce a corresponding revision\\
in subsequent behavior?\\[6pt]
$\Delta E \rightarrow \Delta Z_t \rightarrow \Delta B_{>t}$
}}}

}} 

\end{minipage}

\endgroup

\caption{ARC-AGI as a diagnostic setting for adaptive reasoning.
\textbf{(a)} Across-task adaptivity can be probed by controlled task
variations that preserve or modify the relevant structure, testing whether
behavioral change tracks structural change.
\textbf{(b)} Within-task adaptivity can be probed during interaction by
asking whether new evidence induces corresponding revisions in intermediate
decisions and subsequent behavior, rather than merely triggering additional
search.}
\label{fig:arc-diagnostic-adaptivity}
\end{figure*}

ARC-AGI provides a useful diagnostic case because it spans two related but distinct reasoning settings.
ARC-AGI-1 and ARC-AGI-2 use static input--output tasks, where a solver infers a transformation from sparse demonstrations and applies it to a novel input~\cite{chollet2019arc,chollet2025arc2}.
ARC-AGI-3 instead introduces interactive environments in which an agent acts, observes consequences, and reasons from subsequent evidence~\cite{chollet2026arc3}.
Together, we use them as complementary diagnostics, not alignment benchmarks, for probing abstraction transfer across controlled task variations and evidence-sensitive revision during interaction.

\textbf{Across-task adaptivity in ARC-AGI-1/2.}
In ARC-AGI-1 and ARC-AGI-2, evaluation ultimately emphasizes whether the correct output is produced~\cite{chollet2019arc,chollet2025arc2}.
Yet identical outcomes may arise from different test-time procedures: a solver may infer a compact transformation, search candidate programs, sample and rank hypotheses, refine an earlier proposal, or adapt at test time~\cite{jeon2026neuralmcts,pourcel2025self,akyurek2025ttt}.
Final accuracy does not reveal which capability accounts for success, while resource reporting alone does not establish whether an inferred abstraction transfers when a task changes.

ARC tasks also support controlled variation.
Surface properties such as colors, object identities, positions, orientations, or distractors can change while preserving a relevant relation, whereas the relation itself can change while other properties remain similar~\cite{moskvichev2023conceptarc}.
Such comparisons test whether behavior tracks the structure preserved or changed by an intervention rather than merely succeeding independently on related tasks.

\textbf{Within-task adaptivity in ARC-AGI-3.}
ARC-AGI-3 shifts from static transformation inference to sequential interaction~\cite{chollet2026arc3}.
An agent acts, observes the consequence, and obtains evidence that may support or contradict its current interpretation.
Eventual success, however, may arise from evidence-sensitive revision or simply continued exploration and search.
The relevant distinction is whether subsequent decisions change specifically in response to what earlier interactions revealed.

Relative human action efficiency (RHAE) adds an efficiency perspective by comparing the actions humans and agents require to succeed~\cite{chollet2026arc3}.
Human trajectories can further reveal strategies, failures, and points of revision~\cite{kim2026arctraj,legris2025harc}.
Neither efficient behavior nor trajectory similarity establishes that the same abstraction was acquired or that a hypothesis was revised for the same reason~\cite{kim2025intention}.
They are diagnostic references rather than normative definitions of reasoning.

Fig.~\ref{fig:arc-diagnostic-adaptivity} summarizes how the two scales of adaptivity introduced in Sec.~3 can be instantiated in ARC-AGI.
Across tasks, controlled variations probe abstraction transfer; within a task, interaction probes evidence-sensitive revision.
The next section turns these diagnostic distinctions into concrete intervention-oriented evaluation protocols.
\section{Toward Intervention-Oriented Evaluation}

The preceding analysis suggests evaluating adaptive reasoning through paired behavioral comparisons rather than successful trajectories in isolation.
The aim is to make claims about transfer and revision testable under controlled changes without prescribing a single benchmark or scalar score.
An intervention-oriented evaluation should specify what is changed, what remains fixed, what behavioral response is expected, and what test-time resources are available.

\textbf{Across-task protocol.}
Across-task adaptivity can be evaluated by constructing paired task variants from a base task $\tau$.
Each intervention should identify the task structure $C(\tau)$ relevant to the reasoning claim and distinguish changes intended to preserve that structure from those intended to modify it.
Concept-preserving variants may alter surface properties, object identities, layouts, or other irrelevant context while keeping the target relation fixed.
Concept-changing variants instead modify the relation that should determine the solution while preserving unrelated structure as much as possible.

The system is then evaluated on paired tasks under the same test-time budget $q$.
Adaptivity does not require identical traces or equal accuracy across variants.
Rather, behavior should remain stable or transform predictably when the relevant structure is preserved, and change appropriately when that structure is modified.
Both intervention types are important: invariance alone can arise when a system ignores the manipulated variable, while sensitivity alone can reflect general instability.
Stronger evidence therefore combines robustness to irrelevant variation with responsiveness to relevant structural change.
Use multiple matched variants when possible to reduce dependence on idiosyncratic examples.

Intervention construction itself affects evaluation validity.
A nominally concept-preserving change may alter task difficulty or introduce a new cue, while a concept-changing intervention may modify several relevant properties at once.
Variants should therefore be as localized as possible, with preserved and changed properties stated explicitly~\cite{gardner2020contrastsets,ribeiro2020checklists}.
Matched controls can test whether the same manipulation affects behavior when the altered feature is irrelevant to the target relation.
Reporting the intervention procedure alongside behavior helps distinguish intervention failures from failures of the reasoning system.

\textbf{Within-task protocol.}
Within-task adaptivity requires a paired comparison around a point at which new evidence becomes available.
At step $t$, the evaluator records an observable intermediate decision $Z_t$, such as a hypothesis, proposal, program, plan, or action.
An intervention then changes $E_t$ to $E_t'$ through an additional observation, counterexample, failed prediction, interaction outcome, verifier signal, or corrective feedback, after which the evaluator records subsequent decisions and behavior.
The expected implication of the new evidence for the current decision should be clear, enabling a targeted test of whether the corresponding reasoning component changes appropriately.

The critical question is whether the resulting change is tied to the introduced evidence.
Contradictory evidence should induce revision of the implicated component when warranted, whereas irrelevant or confirming evidence should not induce arbitrary changes.
Such controls distinguish evidence-sensitive revision from generic instability.
The comparison should also separate revision from repeated search: additional independent candidates may improve success, but provide different evidence from using a failure to revise the next proposal.
Holding the test-time budget fixed, or explicitly reporting differences, is therefore necessary for interpretation.

Whenever possible, comparisons should branch from the same pre-intervention history.
One branch receives informative evidence, while matched branches receive no additional evidence, confirming evidence, or an irrelevant perturbation.
Differences in subsequent decisions provide a cleaner test of whether revision is attributable to the intervention rather than stochastic generation or accumulated trajectory differences.
For stochastic systems, repeated runs across random seeds can help distinguish systematic evidence-sensitive updates from isolated changes~\cite{atil2025nondeterministic}.
These responses need not be collapsed into a single score; instead, report them by intervention type and expected behavioral change.

\textbf{Reporting adaptive evidence.}
Report intervention results alongside conventional outcome and process measures rather than replacing them.
Tab.~\ref{tab:reporting} summarizes the minimum information needed to interpret adaptive reasoning claims.

\begin{table}[t]
\centering
\small
\begin{tabular}{p{0.29\columnwidth} p{0.61\columnwidth}}
\toprule
\textbf{Evidence} & \textbf{Minimum reporting} \\
\midrule

Outcome &
Task success pre- and post-intervention. \\

Reasoning artifact &
Relevant trace, hypothesis, proposal, program, action, or other observable intermediate decision. \\

Test-time budget &
Samples, tokens, search steps, refinement rounds, interactions, tools, or other inference-time resources. \\

Across-task &
What structure is preserved or changed, the paired task variants, and the corresponding behavioral response. \\

Within-task &
What evidence is added or modified, the decision before and after the intervention, and subsequent behavior. \\

Human reference &
Action efficiency, trajectory, or strategy information when used as a diagnostic comparison. \\

\bottomrule
\end{tabular}
\caption{Minimum reporting for interpreting intervention-oriented evidence of adaptive reasoning.}
\label{tab:reporting}
\end{table}

These protocols do not determine whether a model possesses a particular internal mechanism.
They instead strengthen evidence for narrower claims: that an abstraction is reused across controlled task changes, or that a decision is revised in a manner linked to newly available evidence.
This distinction matters for alignment-relevant reasoning systems, where robustness under novelty and responsiveness to corrective information matter beyond success on a fixed evaluation instance.
The value of such evidence is therefore not that it certifies alignment, but that it makes specific claims about adaptive behavior more auditable and falsifiable.
\section{Relation to Alignment-Relevant Properties}

Adaptive reasoning is not equivalent to alignment.
A system may transfer abstractions and revise hypotheses effectively while still pursuing undesirable objectives or responding poorly to human values.
Our framework is therefore narrower: across-task transfer and within-task revision provide behavioral evidence for capabilities that matter under novelty, uncertainty, and corrective feedback.

\textbf{Robustness under novelty.}
Across-task adaptivity matters for robustness because deployment conditions rarely reproduce the exact surface properties seen during training or evaluation.
A robust reasoner should preserve behavior when irrelevant properties change while remaining sensitive to changes in task structure that should affect its decision.
Concept-preserving and concept-changing interventions make this distinction explicit.
They provide more targeted evidence than aggregate out-of-distribution performance alone about whether a system's behavior tracks the structure on which its reasoning claim depends.

\textbf{Corrigibility and responsiveness to feedback.}
Within-task adaptivity is related to a system's ability to respond to corrective information.
When new observations, failures, or human feedback undermine an earlier hypothesis, an adaptive system should revise the implicated decision rather than persist or merely search for another successful output.
Corrigibility is broader, encompassing acceptance of human oversight or correction of a system's behavior or objectives~\cite{soares2015corrigibility,hadfieldmenell2017offswitch}.
Our test instead asks whether corrective evidence can systematically alter subsequent reasoning in practice.

Responsiveness alone is insufficient.
A system that changes its behavior indiscriminately in response to irrelevant, noisy, or adversarial feedback may appear adaptive while remaining unreliable.
Alignment-relevant responsiveness depends not only on whether behavior changes, but on whether the change is appropriately tied to the evidence that should matter.
The relevance and specificity criteria introduced in Sec.~3 help distinguish targeted revision from generic sensitivity.

\textbf{Auditability.}
Intervention-oriented evaluation also makes reasoning claims easier to audit.
Outcome accuracy shows whether a system succeeded, while traces and process scores reveal aspects of how it reasoned.
Controlled interventions add a paired comparison between what information changed and how behavior responded.
Combined with explicit reporting of test-time budgets and intermediate decisions, such comparisons make claims about transfer and revision more falsifiable and easier to distinguish from gains due primarily to additional search, sampling, or selection.

These diagnostics should complement, not replace, evaluations of objectives, preferences, policy compliance, or other alignment targets.
This distinction separates adaptive competence from the desirability of the objective being pursued.
Evidence of transfer or revision therefore does not certify that a system is aligned, safe, or causally understood; it provides more specific evidence for alignment-relevant claims about robustness, responsiveness to correction, and auditability under changing conditions.
\section{Discussion}

\textbf{Adaptive evidence complements existing evaluation.}
Our proposal does not imply that final-answer accuracy, reasoning traces, process evaluation, or efficiency measures should be displaced.
Outcome metrics remain necessary for establishing whether a system can solve a task, while traces and process-level assessments expose aspects of the reasoning associated with a solution.
Resource and human-relative measures further reveal computational or behavioral costs that accuracy alone obscures.
The limitation is one of evidential scope: none of these measures by itself establishes whether reasoning transfers across controlled task changes or revises appropriately when relevant evidence changes.
Adaptive evidence should therefore complement, rather than replace, existing evaluation.

\textbf{Behavioral proxies are imperfect.}
The framework does not require a particular reasoning representation.
A system may express intermediate hypotheses through natural language, programs, plans, actions, tool calls, or other observable decisions, while some systems may expose no explicit symbolic hypothesis at all.
The variable $Z_t$ is therefore a system-dependent behavioral proxy rather than a claim that all reasoners share a common intermediate representation.
This flexibility broadens the framework's applicability, but also introduces measurement uncertainty.
An observable proposal or action may only partially reflect the decision process governing subsequent behavior.
Claims about revision should therefore be calibrated to what the chosen observable can support.

\textbf{Behavioral intervention does not identify mechanism.}
Controlled variation strengthens evidence about dependencies between information and behavior, but it does not uniquely identify the mechanism producing them.
Different mechanisms may exhibit similar patterns of transfer or revision, and even localized interventions may alter task difficulty, introduce unintended cues, or affect several properties at once.
Within-task interventions may also change the trajectory in ways unrelated to the intended evidence.
Localized interventions, matched controls, repeated comparisons, and fixed or transparently reported test-time budgets reduce these ambiguities but cannot eliminate them.
Our framework therefore supports stronger behavioral claims about transfer and revision without establishing mechanistic causality.

\textbf{Adaptation must be selective.}
Adaptivity should not be equated with behavioral change itself.
A system that changes its decisions in response to irrelevant, noisy, or adversarial information may appear responsive while remaining unreliable.
Conversely, persistence can be appropriate when new information does not bear on the current hypothesis.
The target is selective adaptation: stability to changes that should not matter and sensitivity to changes that should.
This is why relevance, specificity, and consistency are central to interpreting both across-task and within-task evidence.
This selectivity also suggests asymmetric controls.
Negative controls should alter information irrelevant to the current hypothesis and test for stability, whereas positive controls should alter evidence that warrants revision and test for sensitivity.
Evaluating both responses helps distinguish genuine evidence dependence from generic conservatism or instability.

\textbf{Adaptive evidence should not be collapsed prematurely.}
Across-task transfer and within-task revision capture different failure modes; performance on one does not imply performance on the other.
Concept-preserving and concept-changing interventions test complementary properties: a system may appear robust because it ignores relevant changes, or appear responsive because it changes indiscriminately.
An aggregate adaptivity score could obscure these distinctions.
Reporting results separately by adaptivity scale, intervention type, and expected behavioral response preserves information about where a system succeeds or fails.
Aggregation may still be useful for comparison, but it should follow, not replace, this disaggregated evidence.

\textbf{The framework is not specific to ARC-AGI.}
ARC-AGI serves as a diagnostic case because its tasks make controlled variation, sparse evidence, and interactive feedback comparatively explicit.
The same evaluation logic can apply whenever an evaluator can specify a task property to preserve or change, introduce evidence with a known implication, and observe a subsequent decision or action.
Examples include program synthesis, planning, tool-using agents, and interactive problem solving.
Concrete interventions and behavioral proxies will differ across domains, but the underlying question remains the same: does behavior track the information the claimed reasoning capability should depend on?
This broader scope also means that intervention design must remain domain-specific rather than treating ARC-style transformations as a universal template.

The resulting perspective is intentionally narrower than a complete theory of reasoning.
It asks whether a claimed reasoning capability exhibits predictable behavioral dependencies on the task structure and evidence it should use.
This perspective does not resolve every ambiguity about internal reasoning, but it makes claims about abstraction transfer and hypothesis revision more explicit, testable, and falsifiable.
\section{Conclusion}

Reasoning evaluation now extends beyond final-answer accuracy to reasoning traces, process quality, test-time resources, and human-relative efficiency.
These measures remain valuable, but they do not establish whether reasoning adapts when relevant task structure or available evidence changes.

We characterized adaptive reasoning at two complementary scales.
Across-task adaptivity tests whether acquired abstractions transfer or recombine under controlled task variation, while within-task adaptivity tests whether intermediate decisions revise in response to new observations, failures, or feedback.
Using ARC-AGI as a diagnostic case study, we showed how these distinctions motivate paired intervention-oriented comparisons with explicit reporting of test-time budgets and observable decisions.

Adaptive evidence does not certify alignment or reveal a model's complete internal mechanism.
Its value is narrower but consequential: it makes claims about transfer and revision more directly testable, falsifiable, and auditable.
Reasoning claims require evidence about what a system produces and how its behavior changes when information relevant to its reasoning is deliberately changed.

\bibliography{99_reference}

\newpage
%
%
\providecommand{\appcell}[1]{{\raggedright #1\par}}
\providecommand{\appcenter}[1]{{\centering #1\par}}

\onecolumn

\appendix
\section{Technical Supplement}
\label{sec:supplementary}

This supplementary material provides additional positioning and worked examples for the intervention-oriented evaluation framework introduced in the main paper. The examples illustrate how across-task abstraction transfer and within-task evidence-sensitive revision can be operationalized under explicit test-time resource reporting. They are intended as demonstrations of the evaluation procedure rather than as a new ARC-AGI benchmark or a comprehensive model comparison.
\subsection{Relation to Closely Related Evaluation Paradigms}
\label{sec:supp_related}

Several existing evaluation methods use controlled variations of an input or task to examine how model behavior changes. Contrast sets construct minimally different examples to probe local decision boundaries \citep{suppGardner2020}. CheckList uses behavioral tests designed around specific capabilities rather than aggregate accuracy alone \citep{suppRibeiro2020}. In the ARC domain, ConceptARC groups tasks by human-interpretable concepts and evaluates whether models generalize across tasks that share those concepts \citep{suppMoskvichev2023}. Our across-task evaluation follows a related idea, but focuses on whether behavior changes consistently with the task structure that an intervention preserves or modifies.

Belief revision is closely related to our definition of within-task adaptivity. Belief-R tests whether language models revise a prior conclusion when additional premises warrant a change, and whether they retain the conclusion when the new information does not warrant one \citep{suppWilie2024}. Our within-task evaluation uses the same basic distinction. Contradictory evidence should change the part of a hypothesis that it bears on, while confirming or irrelevant evidence should not lead to unrelated changes. We apply this logic to observable intermediate decisions more broadly, including hypotheses, programs, plans, actions, and tool calls.

Counterfactual interventions have also been used to test whether intermediate reasoning affects later reasoning or final answers \citep{suppXiong2025}. These studies focus on whether changing a reasoning artifact changes subsequent computation. Our evaluation asks a different question: whether a change in task-relevant information produces the corresponding change in reasoning or behavior. These properties need not coincide. A reasoning trace may influence the final answer without being revised when corrective evidence is introduced. Likewise, a different final answer does not by itself show that the part of the hypothesis contradicted by the new evidence was revised.

The novelty of our framework is therefore not the use of controlled interventions itself. We organize adaptive reasoning into two evaluation settings, across-task abstraction transfer and within-task evidence-sensitive revision, and evaluate both by relating controlled changes in task structure or evidence to subsequent behavior. We also make the available test-time resources explicit so that evidence-sensitive revision can be distinguished from gains obtained through additional search or sampling.

\begin{table}[htbp!]
\centering
\small
\setlength{\tabcolsep}{5pt}
\begin{tabular}{
p{2.7cm}
p{3.7cm}
p{4.2cm}
p{5.0cm}}
\toprule
\appcell{\textbf{Approach}}
& \appcell{\textbf{Variation or intervention}}
& \appcell{\textbf{Main evaluative target}}
& \appcell{\textbf{Connection to adaptive reasoning}} \\
\midrule

\appcell{Contrast Sets}
& \appcell{Local changes to an input or task instance}
& \appcell{Sensitivity of model predictions to meaningful local contrasts}
& \appcell{Provides a basis for constructing controlled task variants, which we use to test whether behavior follows preserved or changed task structure.} \\

\appcell{CheckList}
& \appcell{Behavioral test conditions and targeted input perturbations}
& \appcell{Capability-specific behavioral testing}
& \appcell{Motivates targeted tests and controls that isolate the behavior associated with a specified capability.} \\

\appcell{ConceptARC}
& \appcell{Concept-centered families of ARC tasks}
& \appcell{Generalization across tasks that share a concept}
& \appcell{Closely related to across-task adaptivity, which additionally distinguishes concept-preserving from concept-changing interventions.} \\

\appcell{Belief-R}
& \appcell{Newly supplied textual evidence}
& \appcell{Revision or maintenance of a prior conclusion}
& \appcell{Closely related to within-task adaptivity, which applies the same distinction to observable decisions beyond textual conclusions.} \\

\appcell{Thinking-draft faithfulness}
& \appcell{Intermediate reasoning content}
& \appcell{Causal influence of reasoning on later reasoning or answers}
& \appcell{Tests dependence on intervened reasoning content, whereas our within-task evaluation changes task-relevant evidence and examines the resulting revision.} \\

\appcell{\textbf{Our framework}}
& \appcell{Task structure or task-specific evidence}
& \appcell{Abstraction transfer and evidence-sensitive revision}
& \appcell{Evaluates across-task and within-task adaptivity under controlled changes while making test-time resources explicit.} \\

\bottomrule
\end{tabular}
\caption{Relationship to closely related evaluation paradigms. The comparison focuses on the information varied by each approach and the behavioral question it is designed to answer.}
\label{tab:supp_related}
\end{table}
\newpage 

\subsection{Worked Example of Across-Task Adaptivity}
\label{sec:supp_across}

Across-task adaptivity asks whether behavior changes in a way that matches controlled changes in the task structure relevant to a reasoning claim. Let $C(\tau)$ denote the evaluator-specified structure of task $\tau$. For an intervention $I_T$ producing $\tau' = I_T(\tau)$, the main paper defines
\begin{equation*}
\begin{aligned}
\Delta C &= C(\tau') \ominus C(\tau),\\
\Delta B &= B_{\theta}(\tau', E_0; q) \ominus B_{\theta}(\tau, E_0; q).
\end{aligned}
\end{equation*}
The question is whether $\Delta C \sim \Delta B$ under a task-appropriate consistency relation. A concept-preserving intervention should preserve the behavior associated with the target structure, possibly under a predictable transformation, whereas a concept-changing intervention should induce a corresponding change in behavior.

\subsubsection{Downward Relocation in ARC-AGI-1}

Fig.~\ref{fig:across-pulldown} uses three related tasks from ARC-AGI-1, \texttt{1e0a9b12}, \texttt{3618c87e}, and \texttt{25ff71a9}, to illustrate this distinction~\citep{suppARCAGI2026}. These tasks form an illustrative comparison rather than a synthetically controlled intervention set. The evaluator therefore specifies the relevant task structure explicitly before comparing them.

\definecolor{arcblue}{RGB}{30,110,210}
\definecolor{arcred}{RGB}{220,55,55}
\definecolor{arcgreen}{RGB}{45,160,80}
\definecolor{arcyellow}{RGB}{240,195,40}
\definecolor{arcgray}{RGB}{170,170,170}
\definecolor{arcmaroon}{RGB}{153,0,51}
\definecolor{arcmagenta}{RGB}{240,18,190}

\newcommand{\arcgrid}[2]{%
  \fill[black] (0,0) rectangle (#1,#2);
  \draw[step=1cm,gray!55,very thin] (0,0) grid (#1,#2);
  \draw[white!70,line width=.7pt] (0,0) rectangle (#1,#2);
}
\newcommand{\fillcell}[3]{\fill[#3] (#1,#2) rectangle ++(1,1);}
\newcommand{\ioarrow}{\raisebox{0.12cm}{\Large$\rightarrow$}}

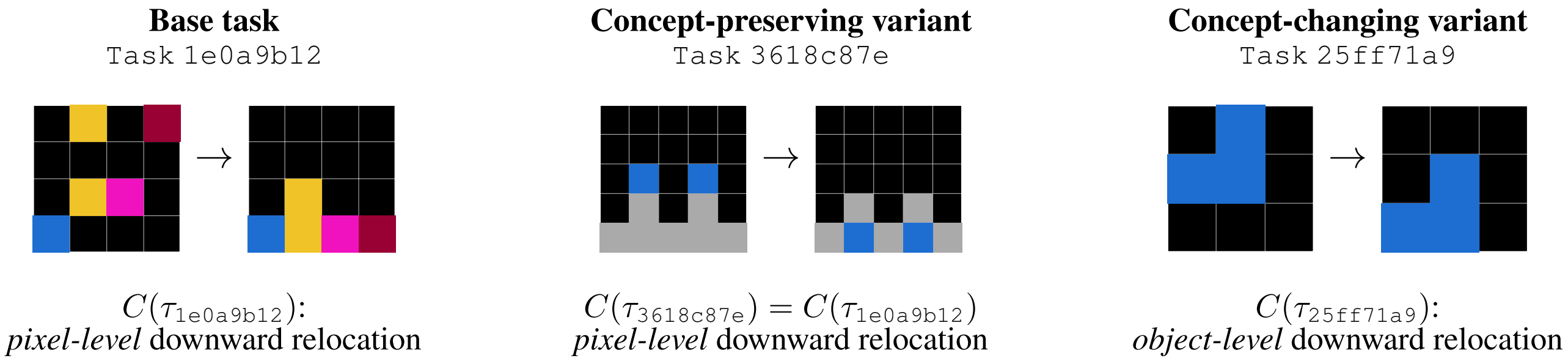
\begin{figure*}[htbp!]
\centering

\begin{subfigure}[t]{0.31\textwidth}
\centering

\parbox[c][0.55cm][c]{\linewidth}{
    \centering
    \textbf{Base task}\\[-1pt]
    {\small \texttt{Task} \texttt{1e0a9b12}}
}

\vspace{3pt}

\begin{minipage}[c][2.25cm][c]{\linewidth}
\centering
\begin{tabular}{c@{\hspace{4pt}}c@{\hspace{4pt}}c}

\begin{tikzpicture}[baseline=(current bounding box.center),scale=.40]
  \arcgrid{4}{4}

  \fillcell{1}{3}{arcyellow}
  \fillcell{3}{3}{arcmaroon}

  \fillcell{1}{1}{arcyellow}
  \fillcell{2}{1}{arcmagenta}

  \fillcell{0}{0}{arcblue}
\end{tikzpicture}
&
\ioarrow
&
\begin{tikzpicture}[baseline=(current bounding box.center),scale=.40]
  \arcgrid{4}{4}

  \fillcell{1}{1}{arcyellow}

  \fillcell{0}{0}{arcblue}
  \fillcell{1}{0}{arcyellow}
  \fillcell{2}{0}{arcmagenta}
  \fillcell{3}{0}{arcmaroon}
\end{tikzpicture}

\end{tabular}
\end{minipage}

\vspace{2pt}

\parbox[t][1.25cm][t]{0.96\linewidth}{
\centering
$C(\tau_{\texttt{1e0a9b12}})$:\\[-1pt]
\textit{pixel-level} downward relocation
}

\end{subfigure}
\hfill
%
\begin{subfigure}[t]{0.31\textwidth}
\centering

\parbox[c][0.55cm][c]{\linewidth}{
    \centering
    \textbf{Concept-preserving variant}\\[-1pt]
    {\small \texttt{Task} \texttt{3618c87e}}
}

\vspace{3pt}

\begin{minipage}[c][2.25cm][c]{\linewidth}
\centering
\begin{tabular}{c@{\hspace{4pt}}c@{\hspace{4pt}}c}

\begin{tikzpicture}[baseline=(current bounding box.center),scale=.32]
  \arcgrid{5}{5}

  \fillcell{1}{2}{arcblue}
  \fillcell{3}{2}{arcblue}

  \fillcell{1}{1}{arcgray}
  \fillcell{3}{1}{arcgray}

  \fillcell{0}{0}{arcgray}
  \fillcell{1}{0}{arcgray}
  \fillcell{2}{0}{arcgray}
  \fillcell{3}{0}{arcgray}
  \fillcell{4}{0}{arcgray}
\end{tikzpicture}
&
\ioarrow
&
\begin{tikzpicture}[baseline=(current bounding box.center),scale=.32]
  \arcgrid{5}{5}

  \fillcell{1}{1}{arcgray}
  \fillcell{3}{1}{arcgray}

  \fillcell{0}{0}{arcgray}
  \fillcell{2}{0}{arcgray}
  \fillcell{4}{0}{arcgray}

  \fillcell{1}{0}{arcblue}
  \fillcell{3}{0}{arcblue}
\end{tikzpicture}

\end{tabular}
\end{minipage}

\vspace{2pt}

\parbox[t][1.25cm][t]{0.96\linewidth}{
\centering
$C(\tau_{\texttt{3618c87e}})=C(\tau_{\texttt{1e0a9b12}})$\\[-1pt]
\textit{pixel-level} downward relocation
}

\end{subfigure}
\hfill
%
\begin{subfigure}[t]{0.31\textwidth}
\centering

\parbox[c][0.55cm][c]{\linewidth}{
    \centering
    \textbf{Concept-changing variant}\\[-1pt]
    {\small \texttt{Task} \texttt{25ff71a9}}
}

\vspace{3pt}

\begin{minipage}[c][2.25cm][c]{\linewidth}
\centering
\begin{tabular}{c@{\hspace{4pt}}c@{\hspace{4pt}}c}

\begin{tikzpicture}[baseline=(current bounding box.center),scale=.53]
  \arcgrid{3}{3}

  \fillcell{1}{2}{arcblue}
  \fillcell{0}{1}{arcblue}
  \fillcell{1}{1}{arcblue}
\end{tikzpicture}
&
\ioarrow
&
\begin{tikzpicture}[baseline=(current bounding box.center),scale=.53]
  \arcgrid{3}{3}

  \fillcell{1}{1}{arcblue}
  \fillcell{0}{0}{arcblue}
  \fillcell{1}{0}{arcblue}
\end{tikzpicture}

\end{tabular}
\end{minipage}

\vspace{2pt}

\parbox[t][1.25cm][t]{0.96\linewidth}{
\centering
$C(\tau_{\texttt{25ff71a9}})$:\\[-1pt]
\textit{object-level} downward relocation
}

\end{subfigure}

\caption{
Across-task adaptivity illustrated with related ARC-AGI-1 tasks.
The base task, \texttt{1e0a9b12}, applies pixel-level downward relocation.
Under the evaluator-specified abstraction, \texttt{3618c87e} is concept-preserving: the manipulation unit and downward-relocation rule remain unchanged despite the introduction of a structured gray background.
In \texttt{25ff71a9}, the manipulation unit changes from individual pixels to a connected object, yielding an object-level downward-relocation rule and a concept-changing variant.
}
\label{fig:across-pulldown}
\end{figure*}

In the base task, \texttt{1e0a9b12}, colored pixels are relocated downward within their columns. We define the structure relevant to this example as \emph{pixel-level downward relocation}. The particular colors and initial positions of the pixels are not part of this abstraction.

Task \texttt{3618c87e} introduces a structured gray background, but the movable elements remain individual pixels and the relevant operation remains downward relocation. Under this evaluator-specified abstraction, we therefore treat
\[
C(\tau_{\texttt{3618c87e}})
=
C(\tau_{\texttt{1e0a9b12}}).
\]
The task serves as a concept-preserving example because the additional background structure changes the visual context without changing the manipulation unit or the downward-relocation rule.

Task \texttt{25ff71a9} retains the general downward-relocation motif but changes the manipulation unit. Connected colored pixels must be treated as a single object whose internal configuration is preserved during relocation. We therefore distinguish its task structure from the pixel-level rule:
\[
C(\tau_{\texttt{25ff71a9}})
\neq
C(\tau_{\texttt{1e0a9b12}}).
\]
This provides a concept-changing example in which the relevant structural difference is the shift from pixel-level to object-level manipulation.

The comparison illustrates why across-task adaptivity is not equivalent to invariance. A solver that has captured the pixel-level abstraction should remain effective when the surrounding background structure changes, as in \texttt{3618c87e}. In contrast, \texttt{25ff71a9} requires behavior appropriate to an object-level representation. The evaluator can therefore distinguish changes that preserve the target abstraction from changes that alter a component of the task structure on which the reasoning claim depends.
\newpage

\subsection{Worked Example of Within-Task Adaptivity}
\label{sec:supp_within}

Within-task adaptivity asks whether new evidence changes an observable intermediate decision $Z_t$ and subsequent behavior. For evidence $E_t$ and an intervention $E'_t=I_E(E_t)$, the main paper considers
\begin{equation*}
\Delta E \longrightarrow \Delta Z_t \longrightarrow \Delta B_{>t}.
\end{equation*}
The key question is whether the change corresponds to what the new evidence supports or rules out.

\subsubsection{Interaction Evidence in ARC-AGI-3}

Consider ARC-AGI-3 task \texttt{r11L}~\citep{suppARCAGI2026}. In Level 1, a purple object must be moved into a target by manipulating two connected endpoints. Clicking a black cell moves the white endpoint, while clicking the dark-gray endpoint swaps the white and dark-gray roles. Light-gray cells are obstacles, and contact between the purple object and an obstacle causes failure.

\begin{figure}[htbp!]
\centering
\small

\begin{minipage}[t]{0.235\textwidth}
\centering
\textbf{(a) Initial state}\\[3pt]
\includegraphics[
    width=\linewidth,
    height=4.3cm
]{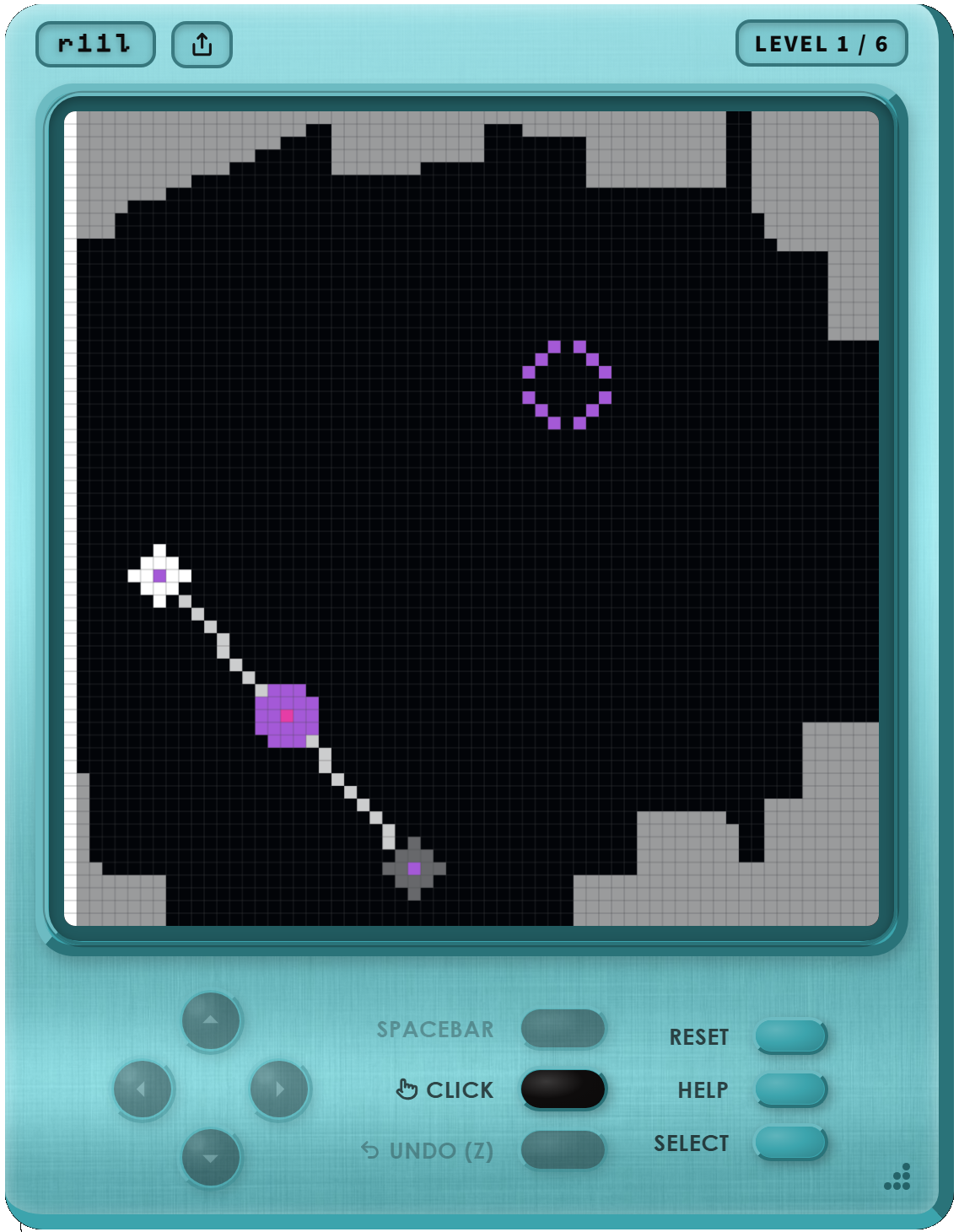}\\[3pt]
Lv.1 begins with the endpoint-control mechanism still to be inferred.
\end{minipage}
\hfill
%
\begin{minipage}[t]{0.235\textwidth}
\centering
\textbf{(b) Confirming evidence}\\[3pt]
\includegraphics[
    width=\linewidth,
    height=4.3cm
]{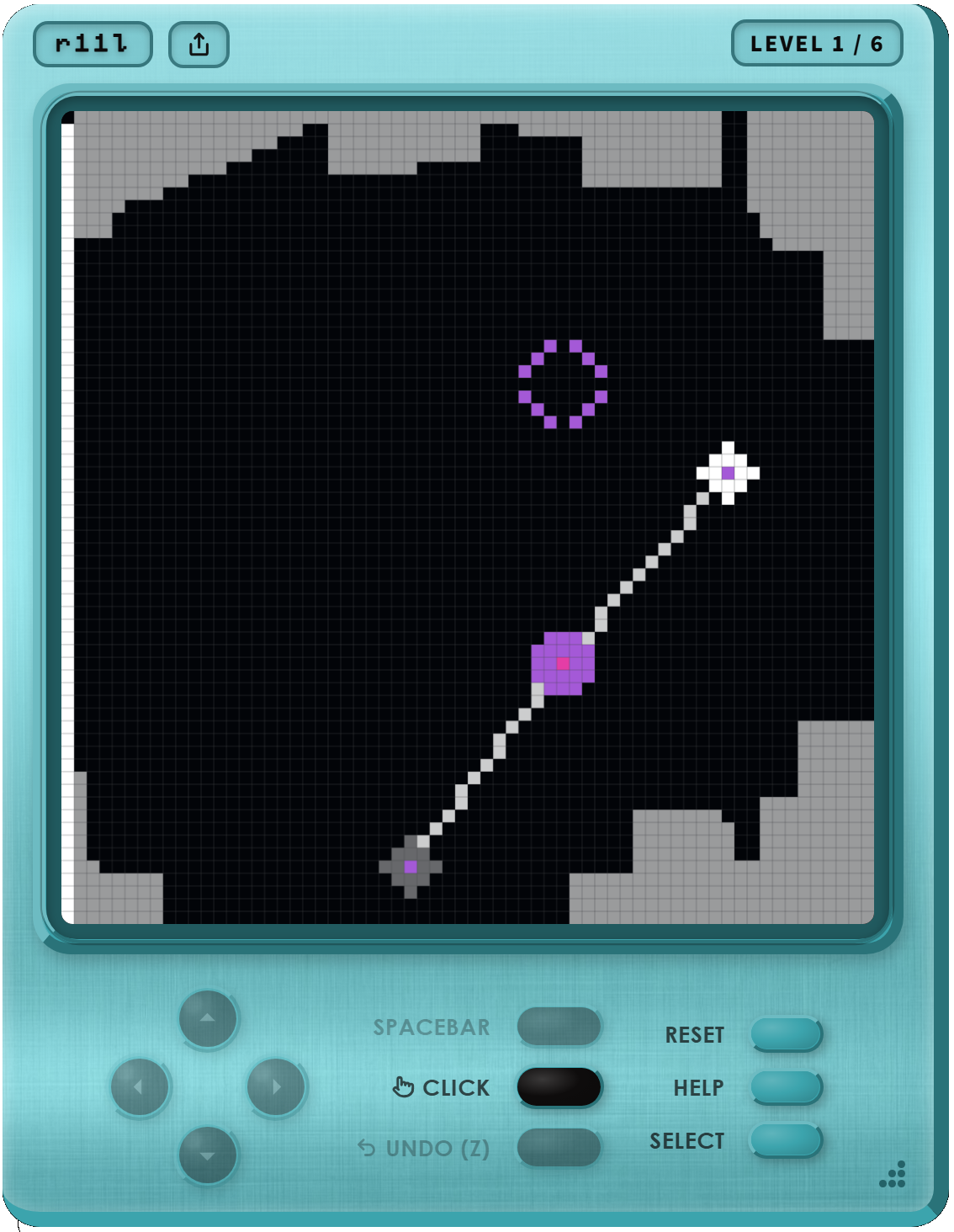}\\[3pt]
Clicking a black cell moves the current white endpoint to the selected location.
\end{minipage}
\hfill
%
\begin{minipage}[t]{0.235\textwidth}
\centering
\textbf{(c) Contradictory evidence}\\[3pt]
\includegraphics[
    width=\linewidth,
    height=4.3cm
]{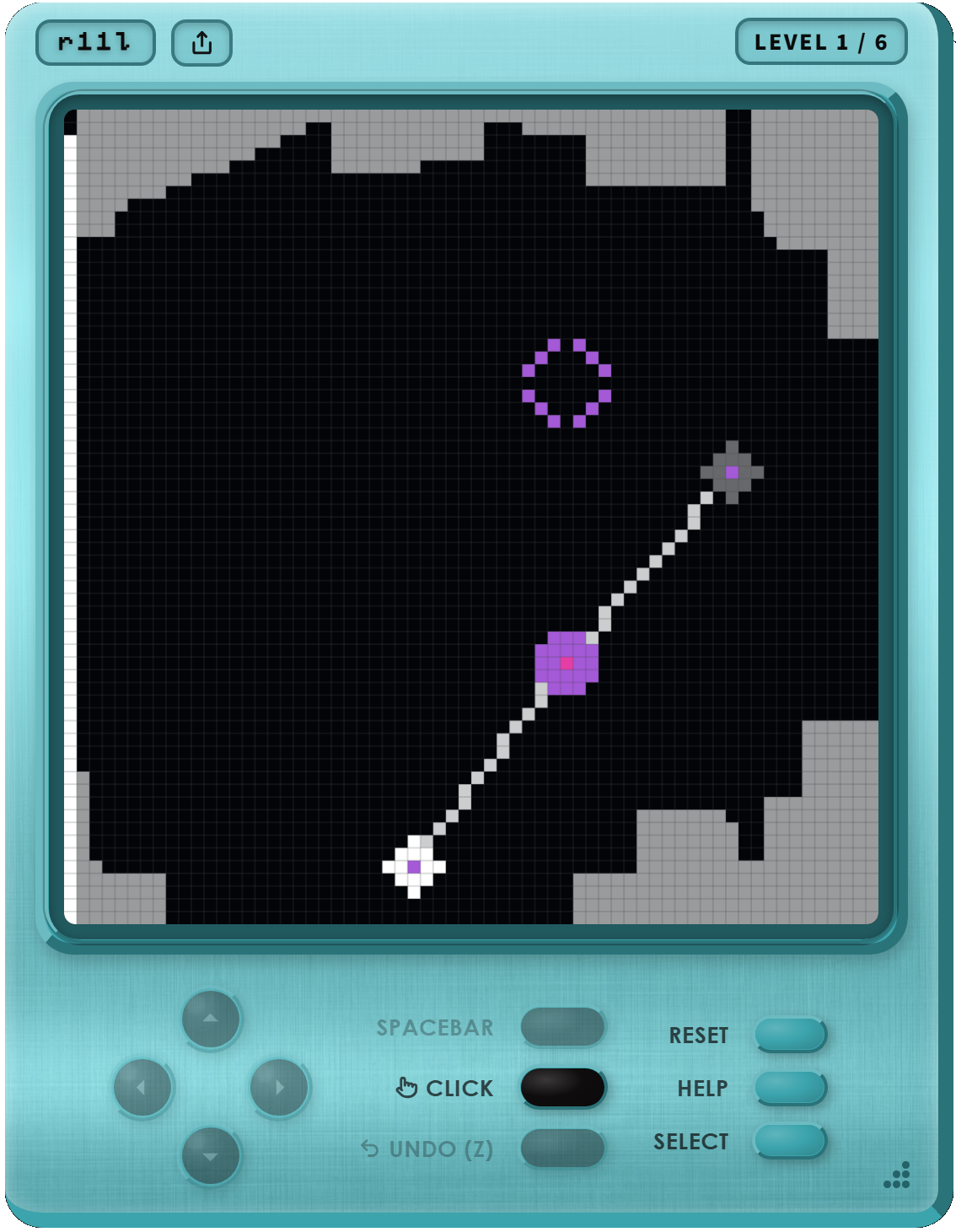}\\[3pt]
Clicking the dark-gray endpoint swaps the white and dark-gray roles.
\end{minipage}
\hfill
%
\begin{minipage}[t]{0.235\textwidth}
\centering
\textbf{(d) Next level}\\[3pt]
\includegraphics[
    width=\linewidth,
    height=4.3cm
]{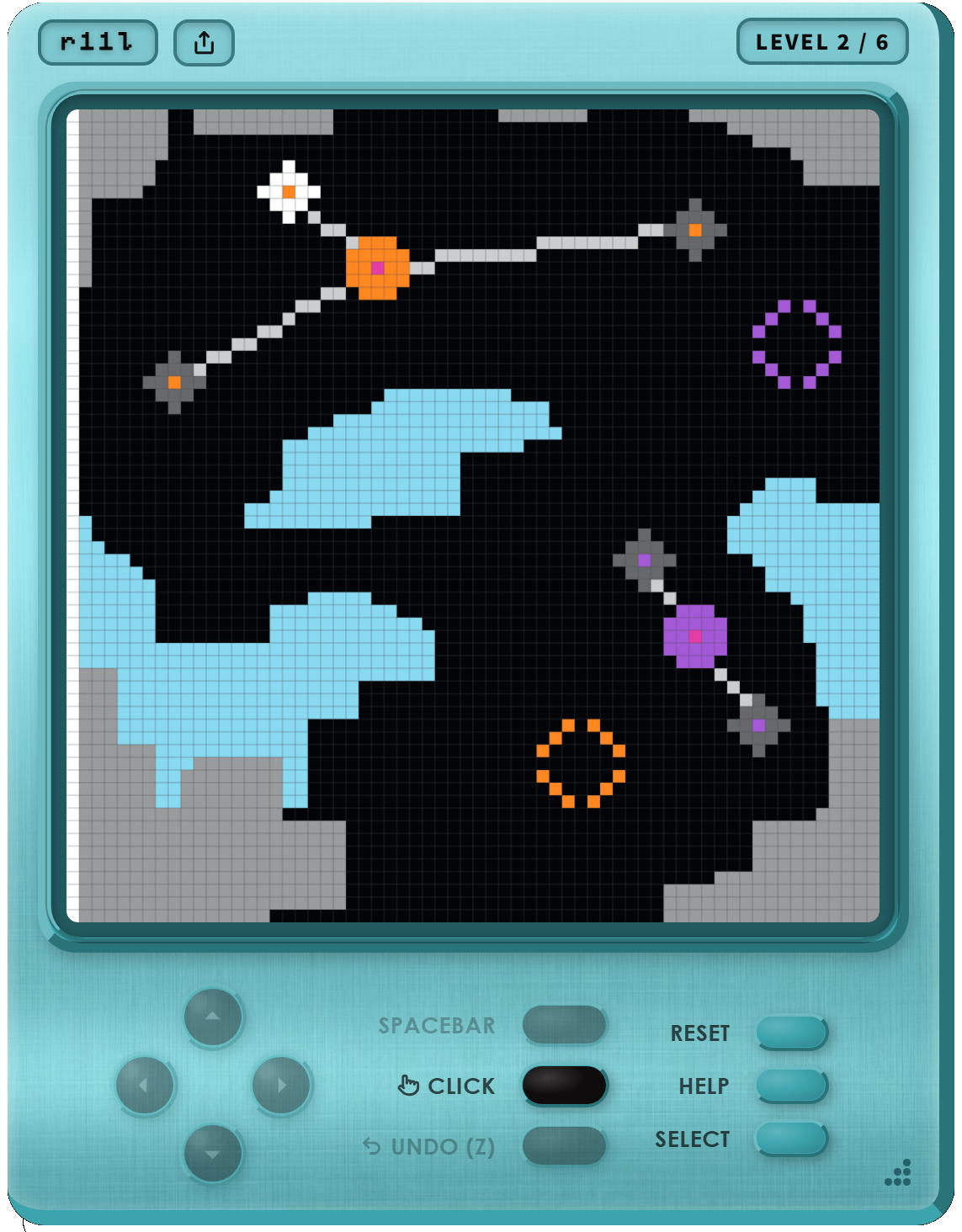}\\[3pt]
Lv.2 introduces new structures while previously learned interaction rules remain relevant.
\end{minipage}

\caption{
Within-task adaptivity in ARC-AGI-3 task \texttt{r11L}.
(a) The initial state of Level 1.
(b) Clicking a black cell moves the current white endpoint, providing confirming evidence about endpoint control.
(c) Clicking the dark-gray endpoint swaps the white and dark-gray roles, contradicting the hypothesis that the dark-gray endpoint is fixed and motivating a targeted revision.
(d) After Level 1 is cleared, Level 2 introduces additional structures while previously acquired interaction rules remain useful.
The sequence illustrates how supported knowledge can be retained while task-specific hypotheses are revised or extended as new evidence becomes available.
}
\label{fig:supp_r11l}
\end{figure}

Fig.~\ref{fig:supp_r11l} shows how interaction reveals these rules and how they remain relevant when new structures appear in Level 2.

Suppose the agent holds $H_1$: \emph{``The white endpoint is movable, whereas the dark-gray endpoint is fixed.''} A black-cell click supports the first component, while a role swap contradicts the second. To test whether the resulting update is evidence-sensitive, the evaluator can branch from the same pre-intervention state and hypothesis, as summarized in Table~\ref{tab:supp_branches}.

\begin{table}[htbp!]
\centering
\footnotesize
\setlength{\tabcolsep}{4pt}
\begin{tabular}{p{2.4cm} p{6.0cm} p{6.3cm}}
\toprule
\textbf{Branch}
& \textbf{Evidence}
& \textbf{Expected response} \\
\midrule

Contradictory
& Clicking the dark-gray endpoint swaps the white and dark-gray roles.
& Revise the claim that the dark-gray endpoint is fixed. \\

Confirming
& Clicking a black cell moves the white endpoint to that cell.
& Retain the claim that the white endpoint is movable. \\

Irrelevant
& Clicking a light-gray obstacle produces no endpoint movement.
& Preserve the endpoint-role hypothesis. \\

No added evidence
& No diagnostic interaction occurs.
& Continue under the same test-time budget. \\

\bottomrule
\end{tabular}
\caption{Same-history branching in ARC-AGI-3 task \texttt{r11L}. Each branch specifies the expected response to different evidence about the current hypothesis.}
\label{tab:supp_branches}
\end{table}

The role-swap observation provides evidence for revision only if the agent updates its endpoint-control hypothesis and subsequent actions reflect the new rule. An agent may instead continue trying different clicks until it succeeds, which demonstrates search rather than evidence-sensitive revision. Evaluation should therefore record $Z_t$ before and after the diagnostic interaction, subsequent behavior, and the test-time budget.
\newpage

\subsection{Operationalizing the Framework}
\label{sec:supp_annotation}

The framework leaves some quantities domain-specific because the relevant task structure and observable behavior depend on the setting. To avoid post-hoc interpretation, the evaluator should specify what is changed, what is preserved, and what response is expected before examining the model's response to the intervention. Tab.~\ref{tab:supp_operationalization} summarizes the information needed for across-task and within-task evaluation.

\begin{table}[htbp!]
\centering
\small
\setlength{\tabcolsep}{4pt}
\begin{tabular}{
p{0.18\columnwidth}
p{0.37\columnwidth}
p{0.37\columnwidth}}
\toprule
\textbf{Stage}
& \textbf{Across-task}
& \textbf{Within-task} \\
\midrule

Before intervention
& Specify the target structure $C(\tau)$ and the properties to be preserved.
& Record the available evidence $E_t$ and an observable decision $Z_t$, such as a hypothesis, program, plan, action, or tool call. \\

Intervention
& Modify a surface property or the target structure while controlling other relevant properties.
& Add or modify evidence to produce $E'_t$ and identify which part of $Z_t$ the new evidence bears on. \\

Expected response
& Specify whether behavior should remain stable, transform predictably, or change with the modified structure.
& Specify whether the implicated decision should be revised or retained given the new evidence. \\

Observed response
& Compare behavior across the paired task variants.
& Compare the decision before and after the intervention and record subsequent behavior $B_{>t}$. \\

Controls
& Use matched variants and report the test-time budget.
& Use no-intervention or irrelevant-evidence branches when possible, and report the test-time budget. \\

\bottomrule
\end{tabular}
\caption{Operationalization of across-task and within-task adaptivity.}
\label{tab:supp_operationalization}
\end{table}

The relevance, specificity, and consistency criteria defined in the main paper are used to interpret these comparisons. Together, they test whether the observed response is tied to the information changed by the intervention and whether similar interventions produce similar response patterns.

For stochastic systems, repeated runs can help separate intervention-related changes from ordinary sampling variation. Test-time resources should be held fixed or reported explicitly so that additional search or sampling is not mistaken for adaptation. Across-task and within-task results should be reported separately because they capture different forms of adaptive behavior.

\subsection{Intervention Validity and Limitations}
\label{sec:supp_validity}

The interpretation of an intervention depends on whether it changes the intended property while leaving other relevant properties unchanged. A concept-preserving intervention may inadvertently alter the target structure, while a concept-changing intervention may modify several relevant factors at once. Interventions should therefore be kept as local as possible, with the changed and preserved properties specified explicitly.

\paragraph{Concept-preserving interventions.}
Consider a task in which color determines the semantic role of an object. Permuting the colors would not be a surface-level change because it could also change the target structure $C(\tau)$. Such a variant should not be treated as a test of invariance to color.

\paragraph{Concept-changing interventions.}
Consider a variant that changes the target spatial relation together with the number, identity, and layout of the objects. If the model's behavior changes, it is difficult to determine which modification caused the response. A more informative comparison changes the target relation while keeping the remaining task properties fixed whenever possible.

\paragraph{Limits of behavioral evidence.}
A consistent pattern of transfer or revision does not identify the internal mechanism that produces it. Different mechanisms may lead to similar observed behavior, and an observable decision $Z_t$ may capture only part of the model's internal reasoning. The resulting evidence therefore supports claims about behavioral transfer and revision, not about a unique internal mechanism.
\subsection{Reporting Checklist}
\label{sec:supp_checklist}

When reporting an intervention-based evaluation, we recommend including:
\begin{enumerate}
    \item the reasoning claim, the relevant task structure or evidence, and what the intervention is intended to change or preserve;
    \item the expected response specified before examining the model output, together with the observed task outcome and intermediate decision when available;
    \item the test-time budget and the controls used to distinguish the effect of the intervention from additional search, sampling, or unrelated variation; and
    \item repeated runs when needed for stochastic systems, with across-task and within-task results reported separately.
\end{enumerate}

These items are intended to make claims about abstraction transfer and evidence-sensitive revision easier to interpret and compare. They do not identify a unique internal reasoning mechanism or establish that a system is aligned.

\begingroup
\renewcommand{\refname}{Supplementary References}

\endgroup

\end{document}